\documentclass{article}

\usepackage[nonatbib, final]{neurips_2019}

\usepackage[utf8]{inputenc} 
\usepackage[T1]{fontenc}    
\usepackage{url}            
\usepackage{booktabs}       
\usepackage{amsfonts}       
\usepackage{nicefrac}       
\usepackage{microtype}      

\usepackage{graphicx}       
\usepackage{amsmath}        
\usepackage{amssymb}        
\usepackage{lettrine}
\usepackage{adjustbox}
\usepackage{multirow}
\usepackage{multicol}
\usepackage{lipsum}
\usepackage[bookmarks=false]{hyperref} 
\hypersetup{
colorlinks,
citecolor=black,
filecolor=black,
linkcolor=black,
urlcolor=black,
}
\usepackage{cleveref}

\usepackage[ruled]{algorithm2e}
\usepackage{subcaption}
\usepackage{comment}
\usepackage{wrapfig}
\usepackage{appendix}

\title{Exploring Self-Supervised Regularization for Supervised and Semi-Supervised Learning}

\author{%
  Phi Vu Tran \\
  Flyreel AI Research \\
  \texttt{vuptran@flyreel.co} \\
}

\begin{document}

\maketitle

\begin{abstract}
Recent advances in semi-supervised learning have shown tremendous potential in overcoming a major barrier to the success of modern machine learning algorithms: access to vast amounts of human-labeled training data. Previous algorithms based on consistency regularization can harness the abundance of unlabeled data to produce impressive results on a number of semi-supervised benchmarks, approaching the performance of strong supervised baselines using only a fraction of the available labeled data. In this work, we challenge the long-standing success of consistency regularization by introducing self-supervised regularization as the basis for combining semantic feature representations from unlabeled data. We perform extensive comparative experiments to demonstrate the effectiveness of self-supervised regularization for supervised and semi-supervised image classification on SVHN, CIFAR-10, and CIFAR-100 benchmark datasets. We present two main results: (1) models augmented with self-supervised regularization significantly improve upon traditional supervised classifiers without the need for unlabeled data; (2) together with unlabeled data, our models yield semi-supervised performance competitive with, and in many cases exceeding, prior state-of-the-art consistency baselines. Lastly, our models have the practical utility of being efficiently trained end-to-end and require no additional hyper-parameters to tune for optimal performance beyond the standard set for training neural networks. Reference code and data are available at \url{https://github.com/vuptran/sesemi}.
\end{abstract}

\section{Introduction}
Contemporary approaches to supervised representation learning, such as the convolutional neural network (CNN), continue to push the boundaries of research across a number of domains including speech recognition, visual understanding, and language modeling. However, such progresses usually require massive amounts of human-labeled training data. The process of collecting, curating, and hand-labeling large amounts of training data is often tedious, time-consuming, and costly to scale. Thus, there is a growing body of research dedicated to learning with limited labels, enabling machines to do more with less human supervision, in order to fully harness the benefits of deep learning in real-world settings. Such emerging research directions include domain adaptation \cite{da}, low-shot learning \cite{zero-few}, self-supervised learning \cite{revisit-ssl}, and multi-task learning \cite{mtl}. In this work, we propose to combine multiple modes of supervision on sources of labeled and unlabeled data for enhanced learning and generalization. Specifically, our work falls within the framework of semi-supervised learning (SSL) \cite{ssl}, in the context of image classification, which can leverage abundant unlabeled data to significantly improve upon supervised classifiers in the limited labeled data setting. Indeed, in many cases, state-of-the-art semi-supervised algorithms have been shown to approach the performance of strong supervised baselines using only a fraction of the available labeled data \cite{vat2,mean-teacher,self-gan2}.

Our approach to SSL belongs to a class of methods that produce proxy, or surrogate, labels from unlabeled data without requiring human annotations, which are used as targets together with labeled data. Although proxy labels may not reflect the ground truth, they provide surprisingly strong supervisory signals for learning the underlying structure of the data manifold. The training protocol for this class of SSL algorithms simply imposes an additional loss term to the overall objective function of an otherwise supervised algorithm. The auxiliary loss describes the contribution of unlabeled data and is referred to as the \emph{unsupervised} loss component.

\begin{figure}[t]
    \centering
    \begin{minipage}{0.6\textwidth}
        \begin{subfigure}[t]{1.23\textwidth}
            \centering
            \includegraphics[height=1.1in]{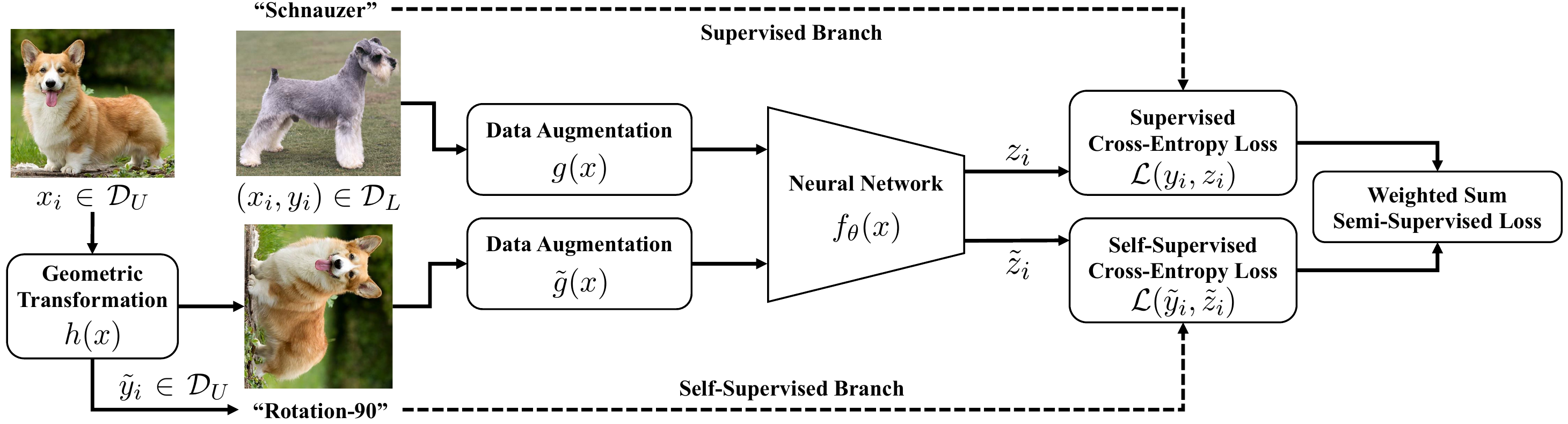}
            \caption{SESEMI architecture for supervised and semi-supervised image classification, with the self-supervised task of recognizing geometric transformations. The function $h(x)$ produces six proxy labels defined as image rotations belonging in the set of $\{0,90,180,270\}$ degrees along with horizontal (left-right) and vertical (up-down) flips.}
            \label{figure1a}
        \end{subfigure}
    \end{minipage}%
    \begin{minipage}{0.8\textwidth}
        \begin{subfigure}[t]{0.7\textwidth}
            \centering
            \includegraphics[height=0.95in]{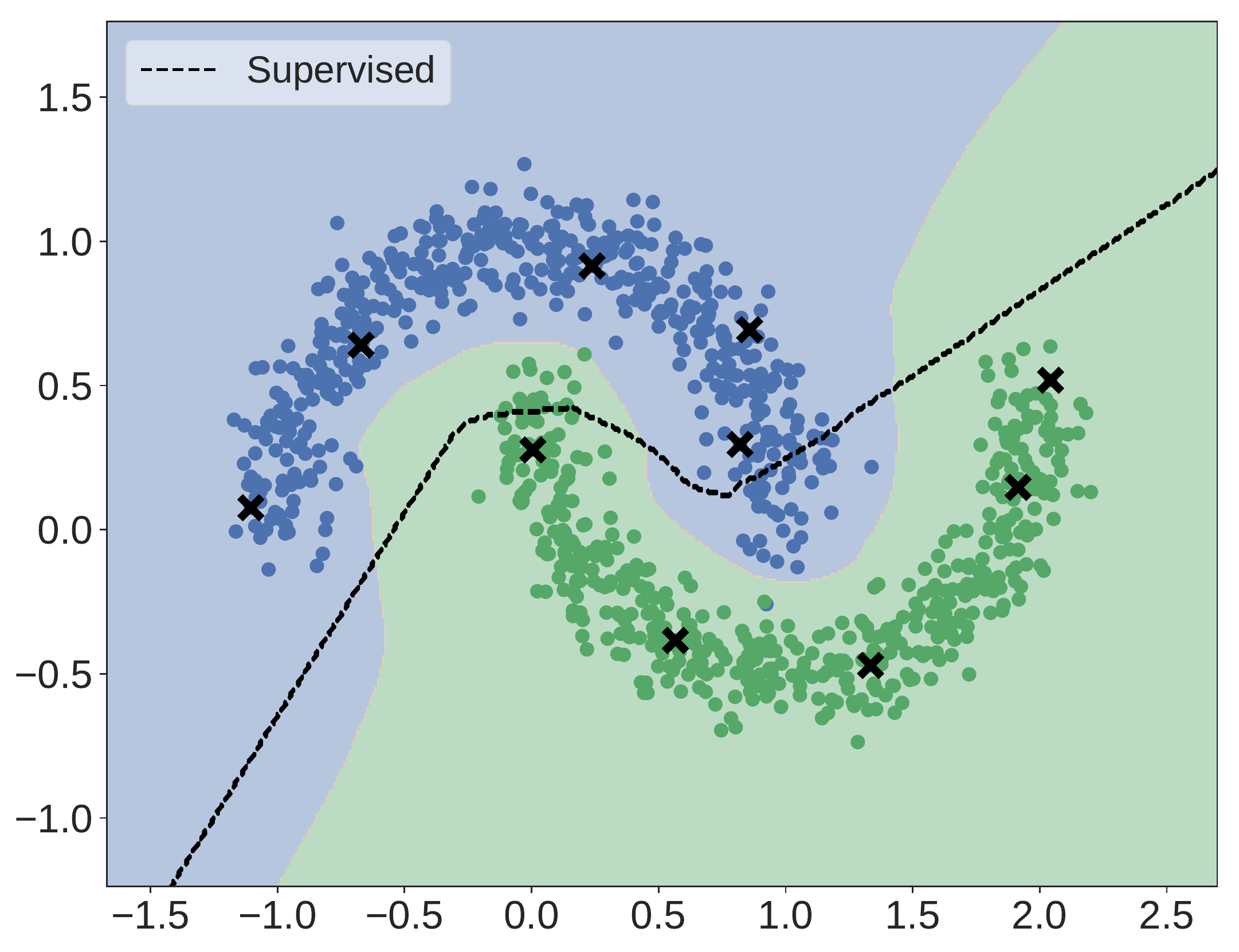}
            \caption{Two moons.}
            \label{figure1b}
        \end{subfigure}
        \begin{subfigure}[t]{0.705\textwidth}
            \centering
            \includegraphics[height=0.95in]{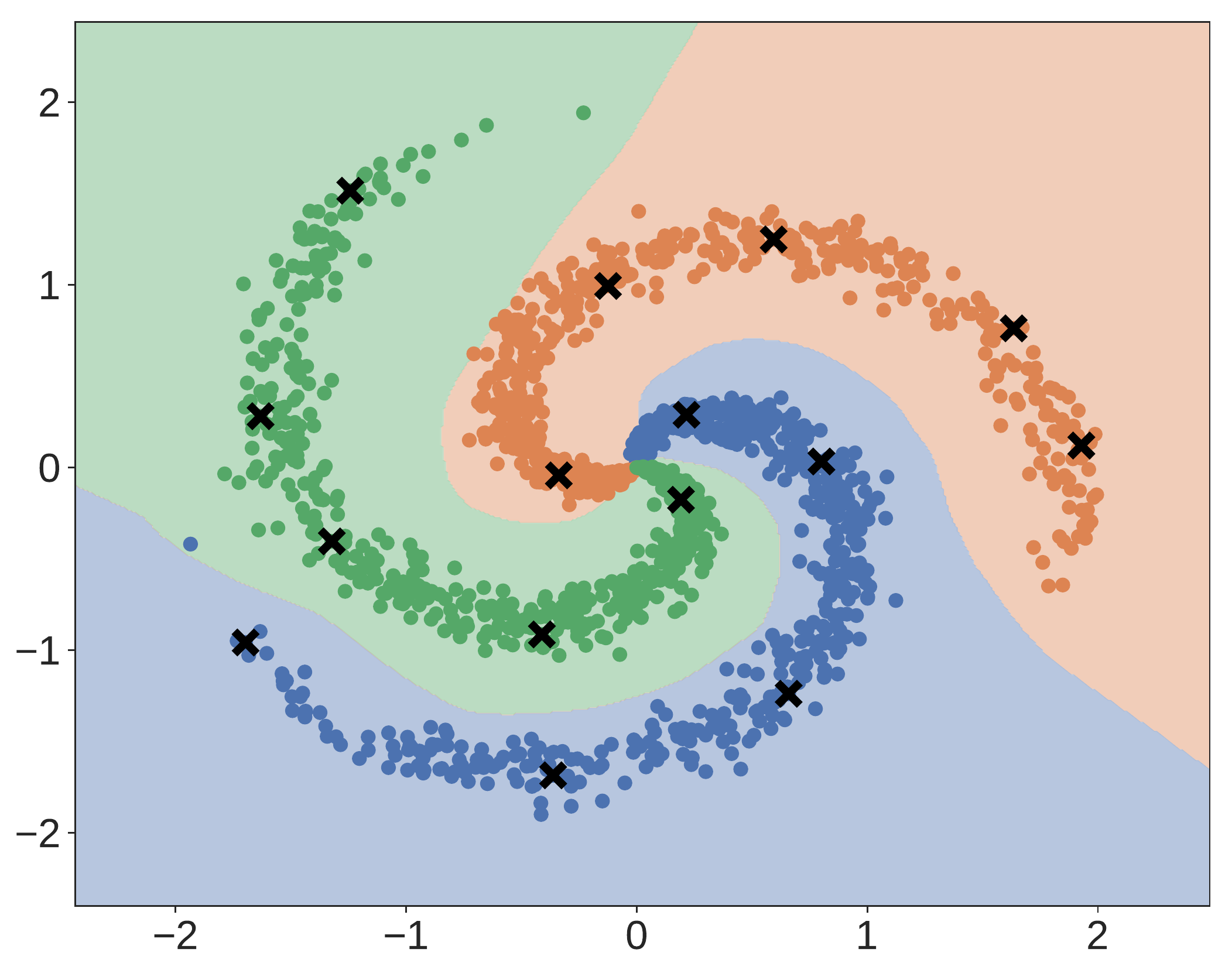}
            \caption{Three spirals.}
            \label{figure1c}
        \end{subfigure}
    \end{minipage}%
    \caption{\textbf{Left (a)} -- The SESEMI architecture. \textbf{Right} -- Demonstration of the SESEMI algorithm on \textbf{(b)} \texttt{two moons} and \textbf{(c)} \texttt{three spirals} synthetic datasets. Labeled examples are marked with the black cross. Decision boundaries separate classes by colors.}
    \label{figure1}
\end{figure}

\subsection{Related Work}
Too many methods have been proposed to learn effective representations from abundant unlabeled data to augment limited ground truth labels for SSL. We summarize three particular categories most related to this work: consistency regularization, adversarial training, and self-supervised learning.

\textbf{Consistency Regularization} ~ Models belonging to this class of SSL algorithms assume a dual role. On one hand, the model learns from labeled data in the conventional supervised manner; simultaneously, the model generates proxy targets for unlabeled data to be learned in conjunction with ground truth labels. During model training, sources of randomness such as dropout \cite{dropout} and stochastic data augmentation, along with varying levels of Gaussian noise in the input data can produce drastically different output predictions. The objective is to stabilize ensembles of predictions by minimizing their mean squared error over randomly perturbed and augmented training examples. The motivation behind this approach is to further regularize the model through the \emph{consistency} principle that perturbations in the input data and/or data augmentation techniques should not significantly change the output of the model \cite{consistency}. Models predicated on the consistency principle, such as Pseudo-Ensembles \cite{pea}, $\Pi$ model \cite{consistency}, Temporal Ensembling \cite{tempens} and Mean Teacher \cite{mean-teacher}, produce stable predictions on unlabeled data, which are used as unsupervised targets. The auxiliary, unsupervised consistency loss term is thus formulated as a regularizer to be jointly trained with the supervised objective on both unlabeled and labeled data for SSL.

\textbf{Adversarial Training} ~ Rather than relying on the model to randomly perturb the input data by way of dropout or data augmentation, Goodfellow et al. (2015) \cite{goodfellow-2015} proposed the concept of adversarial training to approximate the perturbations in the direction that would significantly alter the output of the model. While adversarial training requires access to ground truth labels to perform adversarial perturbations, the Virtual Adversarial Training (VAT) mechanism proposed by Miyato et al. (2017) \cite{vat2} can be applied to unlabeled data to produce an auxiliary unsupervised loss term that is compatible with the consistency regularization framework for SSL. Adversarial training is closely related to generative adversarial networks (GANs) \cite{gans}, which have been proposed for SSL with promising results \cite{cat-gan,ssl-gans2}. Most recently, the self-supervised GANs with auxiliary rotation loss \cite{self-gan2} have been shown to synthesize diverse images at high resolution using only a fraction of the available labels.

\textbf{Self-Supervised Learning} ~ Self-supervised learning is similar in flavor to unsupervised learning, where the goal is to learn rich visual representations from large-scale unlabeled images or videos without using any human annotations. Self-supervised representations are learned by first defining a pretext task, an objective function, for the model to solve and then producing proxy labels to guide the pretext task based solely on the visual information present in unlabeled data. The simplest self-supervised task is minimizing reconstruction error in autoencoders \cite{autoencoder} to learn low-dimensional feature representations, where the proxy labels are the values of the image pixels. More sophisticated self-supervised tasks such as image inpainting \cite{inpainting}, colorizing grayscale images \cite{colorization}, and predicting image rotations \cite{rotations} have shown impressive results for unsupervised visual feature learning. Previous methods utilizing self-supervision for SSL did not combine an auxiliary loss with the supervised objective function but instead followed a two-stage approach: (1) pre-train a self-supervised model on a pretext task to learn useful features from unlabeled data; and (2) transfer the learned representations to downstream applications, via supervised fine-tuning, where labeled training data is scarce.

\subsection{Summary of Contributions}
We introduce a new algorithm to jointly train a \emph{self-supervised loss} term with the traditional supervised objective for the multi-task learning of both unlabeled and labeled data in a single stage. Our work is in direct contrast to prior SSL approaches based on unsupervised or self-supervised learning \cite{ali,rotations}, which require the sequential combination of unsupervised or self-supervised pre-training followed by supervised fine-tuning.

Our approach to utilize the self-supervised loss term both as a regularizer (applied to labeled data) and SSL method (applied to unlabeled data) is analogous to consistency regularization. Although leading approaches based on consistency regularization achieve state-of-the-art SSL results, these methods require careful tuning of many hyper-parameters and are generally not easy to implement in practice. Striving for simplicity and pragmatism, our models with self-supervised regularization require no additional hyper-parameters to tune for optimal performance beyond the standard set for training neural networks. Our work is among the first to challenge the long-standing success of consistency regularization for SSL.

We conduct extensive comparative experiments to validate the effectiveness of our models by showing semi-supervised results competitive with, and in many cases surpassing, previous state-of-the-art consistency baselines. We also demonstrate that supervised learning augmented with self-supervised regularization is a viable and attractive alternative to transfer learning without the need to pre-train a separate model on large labeled datasets. Lastly, we perform an ablation study showing our proposed algorithm is the best among a family of self-supervised regularization techniques, when comparing their relative contributions to SSL performance.

\section{Learning with Self-Supervised Regularization}
We present SESEMI, a conceptually simple yet effective algorithm for enhancing supervised and semi-supervised image classification via self-supervision. The design of SESEMI is depicted in \Cref{figure1a}. The input to SESEMI is a training set of input-target pairs $(x, y) \in \mathcal{D}_L$ and (optional) unlabeled inputs $x \in \mathcal{D}_U$. Typically, we assume $\mathcal{D}_L$ and $\mathcal{D}_U$ are sampled from the same distribution $p(x)$, in which case $\mathcal{D}_L$ is a labeled subset of $\mathcal{D}_U$. However, that assumption may not necessarily hold true in real-world settings where there exists the potential for class-distribution mismatch \cite{ssl-eval}. That is, $\mathcal{D}_L$ is sampled from $p(x)$ but $\mathcal{D}_U$ may be sampled from a different, although somewhat related, distribution $q(x)$. The goal of SESEMI is to train a prediction function $f_\theta(x)$, parametrized by $\theta$, that utilizes a combination of $\mathcal{D}_L$ and $\mathcal{D}_U$ to obtain significantly better predictive performance than what would be achieved by using $\mathcal{D}_L$ alone.

\Cref{figure1b,figure1c} illustrate the SESEMI algorithm on the \texttt{two moons} and \texttt{three spirals} synthetic datasets. Each dataset has 500 examples per class and a label rate of 0.01 (\emph{i.e.,} only 5 examples are labeled per class). The prediction function $f_\theta(x)$ is a multi-layer perceptron with three hidden layers, each hidden layer containing 100 leaky ReLU units \cite{lrelu} with $\alpha=0.1$. We observe that the supervised model is unable to fully capture the underlying shape of the data manifold of \texttt{two moons} when trained only on the labeled examples. Together with unlabeled data, SESEMI is able to learn better decision boundaries of both \texttt{two moons} and \texttt{three spirals} that would result in fewer mis-classifications on the test set. This demonstration illustrates the applicability of SESEMI to other data modalities besides image.

\subsection{Convolutional Architectures}
In principle, the prediction function $f_\theta(x)$ could be any classifier. For comparison and analysis with previous work, we experiment with three high-performance CNN architectures: (i) the 13-layer max-pooling Network-in-Network (NiN) \cite{rotations}; (ii) the 13-layer max-pooling ConvNet \cite{tempens,mean-teacher,vat2,vadd,sntg}; and (iii) the more modern wide residual network with depth 28 and width 2 (WRN-28-2) \cite{wrn,ssl-eval}.

We faithfully follow the original specifications of the NiN, WRN, and ConvNet architectures, so we refer to their papers for details. All architectures have convolutional layers followed by batch normalization \cite{batchnorm} and ReLU non-linearity \cite{relu}, except the ConvNet architecture uses leaky ReLU \cite{lrelu} with $\alpha = 0.1$. The NiN, WRN, ConvNet architectures have roughly 1.01, 1.47, and 3.13 million parameters, respectively.

For both supervised and semi-supervised settings, we separate the input data into labeled and unlabeled branches, and apply the same CNN model to both. Note that the unlabeled branch consists of all available training examples, but without ground truth label information. One can view SESEMI as a multi-task architecture that has a common CNN ``backbone'' to learn a shared representation of both labeled and unlabeled data, and an output ``head'' for each task. The ConvNet backbone computes an abstract $6 \times 6 \times 128$ dimensional feature representation from the input image, while the NiN and WRN backbones have an output of $8 \times 8 \times 192$ and $8 \times 8 \times 128$ dimensions, respectively. Each task has extra layers in the head, which may have a complex structure, and computes a separate loss. The head of the labeled branch has a global average pooling layer followed by softmax activation to evaluate the supervised task with standard categorical cross-entropy loss. For the unlabeled branch, we define a self-supervised pretext task to be learned in conjunction with the labeled branch.

\begin{algorithm}[t]
{\caption{SESEMI mini-batch training.}\label{algorithm1}}
\DontPrintSemicolon
\SetAlgoNoLine
\SetKwInOut{Require}{Require}
\Require{~~Training set of labeled input-target pairs $(x, y) \in \mathcal{D}_L$. \\
         ~~Training set of unlabeled inputs $x \in \mathcal{D}_U$. \\
         ~~Geometric transformation function $h(x)$ producing proxy labels $\tilde{y} \in \mathcal{D}_U$. \\
         ~~Input data augmentation functions $g(x)$ and $\tilde{g}(x)$. \\
         ~~Neural network architecture $f_\theta(x)$ with trainable parameters $\theta$.
}
\BlankLine
\For{each epoch over $\mathcal{D}_U$}{ 
    {\setlength{\abovedisplayskip}{0pt}%
    \begin{flalign*}
        B_L &\leftarrow g\left(x_{i \in \mathcal{D}_L}\right) && \triangleright ~ \text{Sample mini-batches of augmented labeled inputs.} &\\
        B_U &\leftarrow \tilde{g}\left(h\left(x_{i \in \mathcal{D}_U}\right)\right) && \triangleright ~ \text{Sample mini-batches of augmented unlabeled inputs.}
    \end{flalign*}
    }
    \hspace{-3pt}\For{each mini-batch}{ 
      {\setlength{\abovedisplayskip}{0pt}%
      \begin{flalign*}
        &z_{i \in B_L} \leftarrow f_\theta\left(B_L\right) && \triangleright ~ \text{Compute model outputs for labeled inputs.} \\
        &\tilde{z}_{i \in B_U} \leftarrow f_\theta\left(B_U\right) && \triangleright ~ \text{Compute model outputs for unlabeled inputs.} \\
        &\mathcal{L} \leftarrow - \textstyle{\frac{1}{|B_L|}\sum_{i \in B_L}\sum_{c \in C}y_{ic} \log (z_{ic})} && \triangleright ~ \text{Supervised cross-entropy loss.} \\
         &\qquad - \textstyle{\frac{1}{|B_U|}\sum_{i \in B_U}\sum_{k \in K}\tilde{y}_{ik} \log (\tilde{z}_{ik})} && \triangleright ~ \text{Self-supervised cross-entropy loss.} \\
        &~\theta \leftarrow \theta - \nabla_\theta\mathcal{L} && \triangleright ~ \text{Update parameters via gradient descent.}
      \end{flalign*}
      \vspace*{-10pt}
      }
    }
}
\Return{$f_\theta(x)$}
\end{algorithm}

\subsection{Recognizing Image Rotations and Flips as Self-Supervision}
Following \cite{rotations}, we apply a set of discrete geometric transformations on the input image and train the network to recognize the resulting transformations as the self-supervised task. The network architecture for the self-supervised task shares the same CNN backbone with its supervised counterpart but has a separate output head consisting of a global average pooling layer followed by softmax activation. In their original work on self-supervised rotation recognition, Gidaris et al. \cite{rotations} defined the proxy labels to be image rotations belonging in the set of $\{0, 90, 180, 270\}$ degrees, resulting in a four-way classification task. Their models performed well on the rotation recognition task by learning salient visual features depicted in the image, such as location of objects, type, and pose. In this work, we extend the geometric transformations to include horizontal (left-right) and vertical (up-down) flips, resulting in the self-supervised cross-entropy loss over six classes. Further, we propose to train SESEMI on both labeled and unlabeled data simultaneously, which is more efficient and yields better performance than the approach of Gidaris et al. based on the sequential combination of self-supervised pre-training on unlabeled data followed by supervised fine-tuning on labeled data.

\subsection{Integrating Self-Supervised Loss as Regularization}
The algorithmic overview of SESEMI is provided in Algorithm~\ref{algorithm1}. At each training step, we sample two mini-batches having the same number of labeled and unlabeled examples as inputs to a shared CNN backbone $f_\theta(x)$. Note that in a typical semi-supervised setting, labeled examples will repeat in a mini-batch because the number of unlabeled examples is much greater. We forward propagate $f_\theta(x)$ twice, once on the labeled branch $x_{i\in\mathcal{D}_L}$ and another pass on the unlabeled branch $x_{i\in\mathcal{D}_U}$, resulting in softmax prediction vectors $z_i$ and $\tilde{z}_i$, respectively. We compute the supervised cross-entropy loss $\mathcal{L}_\textsc{super}(y_i, z_i)$ using ground truth labels $y_i$ and compute the self-supervised cross-entropy loss $\mathcal{L}_\textsc{self}(\tilde{y}_i, \tilde{z}_i)$ using proxy labels $\tilde{y}_i$ generated from image rotations and flips. The parameters $\theta$ are learned via backpropagation by minimizing the multi-task SESEMI objective function defined as the weighted sum of supervised and self-supervised loss components:
\begin{equation*}
\mathcal{L}_\textsc{sesemi} = \mathcal{L}_\textsc{super}(y_i, z_i) + w\mathcal{L}_\textsc{self}(\tilde{y}_i, \tilde{z}_i).
\end{equation*}
Our formulation of the SESEMI objective treats the self-supervised loss as a regularization term, and $w > 0$ is the regularization hyper-parameter that controls the relative contribution of self-supervision in the overall objective function.

In previous SSL approaches based on consistency regularization, such as $\Pi$ model and Mean Teacher, $w$ was formulated as the consistency coefficient and was subjected to considerable tuning, on a per-dataset basis, for optimal performance. We experimented with different values for the weighting parameter $w$ in SESEMI and found $w = 1$ yields consistent results across all datasets and CNN architectures, suggesting that supervised and self-supervised losses are relatively balanced and compatible for image classification. Moreover, setting $w =1$ leads to a convenient benefit of having one less hyper-parameter to tune. We backpropagate gradients to both branches of the network to update $\theta$, similar to $\Pi$ model. The self-supervised loss term has a dual purpose. First, it enables SESEMI to learn additional, complementary visual features from unlabeled data that help guide its decision boundaries along the data manifold. Second, it is compatible with conventional supervised learning without unlabeled data by serving as a strong regularizer against geometric transformations for improved generalization. We refer to SESEMI models trained with self-supervised regularization on labeled data  as \emph{augmented supervised learning} (ASL). At inference time, we simply take the supervised branch of the network to make predictions on test data and discard the self-supervised branch.

\section{Empirical Evaluation}
We follow standard evaluation protocol for SSL, in which we randomly sample varying fractions of the training data as labeled examples while treating the entire training set, discarding all label information, as the source of unlabeled data. We train a model with both labeled and unlabeled data according to SESEMI (Algorithm~\ref{algorithm1}) and compare its performance to that of the same model trained using only the labeled portion in the traditional supervised manner. For ASL, we train SESEMI using the ConvNet architecture on labeled data, but augment the supervised objective with self-supervised regularization. The performance metric is classification error rate. We expect a good SSL algorithm to yield better results (lower error rate) when unlabeled data is used together with labeled data. We closely follow the experimental protocols described in \cite{rotations,tempens,mean-teacher,ssl-eval} to remain consistent with previous work.

\subsection{Datasets and Baselines}
We evaluate our proposed SESEMI algorithm on three benchmark datasets for supervised and semi-supervised image classification: Street View House Numbers (SVHN) \cite{svhn}, CIFAR-10 and CIFAR-100 \cite{cifar}. For details on the datasets and implementation, see \Cref{datasets,implementation}. We also use two auxiliary datasets to augment our experiments on supervised and semi-supervised learning: 80 million Tiny Images \cite{tiny-images} and ImageNet-32 \cite{tiny-imagenet}. Tiny Images is the superset of CIFAR-10 and CIFAR-100 organized into 75,062 generic scene and object categories. ImageNet-32 is the full ImageNet dataset \cite{imagenet} down-sampled to $32 \times 32$ pixels. We use ImageNet-32 for supervised transfer learning experiments. We use Tiny Images as a source of unlabeled extra data to augment SSL on CIFAR-100 and to evaluate SESEMI under the condition of class-distribution mismatch.

We empirically compare our SESEMI models trained with self-supervised regularization against two state-of-the-art baselines for supervised and semi-supervised learning: (a) the RotNet models of Gidaris et al. (2018) \cite{rotations}, which were pre-trained on unlabeled data with self-supervised rotation loss followed by a separate step of supervised fine-tuning on labeled data; and (b) models jointly trained on both unlabeled and labeled data using consistency regularization as the unsupervised loss, namely $\Pi$ model and its Temporal Ensembling (TempEns) variant \cite{tempens}, VAT \cite{vat2}, and Mean Teacher \cite{mean-teacher}.

The RotNet baseline uses the 13-layer max-pooling NiN architecture, whereas the consistency models use the 13-layer max-pooling ConvNet architecture. We also provide a comparison of SESEMI within the unified evaluation framework of Oliver et al. (2018) \cite{ssl-eval}, in which they re-implemented the consistency models using the WRN-28-2 architecture. Thus, our experiments report results from both ConvNet and WRN backbones to evaluate the relative impact of alternative convolutional architectures on SSL performance.

\subsection{Results and Analysis}
\subsubsection{Self-Supervised Regularization Outperforms Pre-Training $+$ Fine-Tuning on CIFAR-10}

\begin{figure}[t]
    \centering
    \begin{minipage}[t]{0.49\textwidth}
        \centering
        \includegraphics[width=\textwidth]{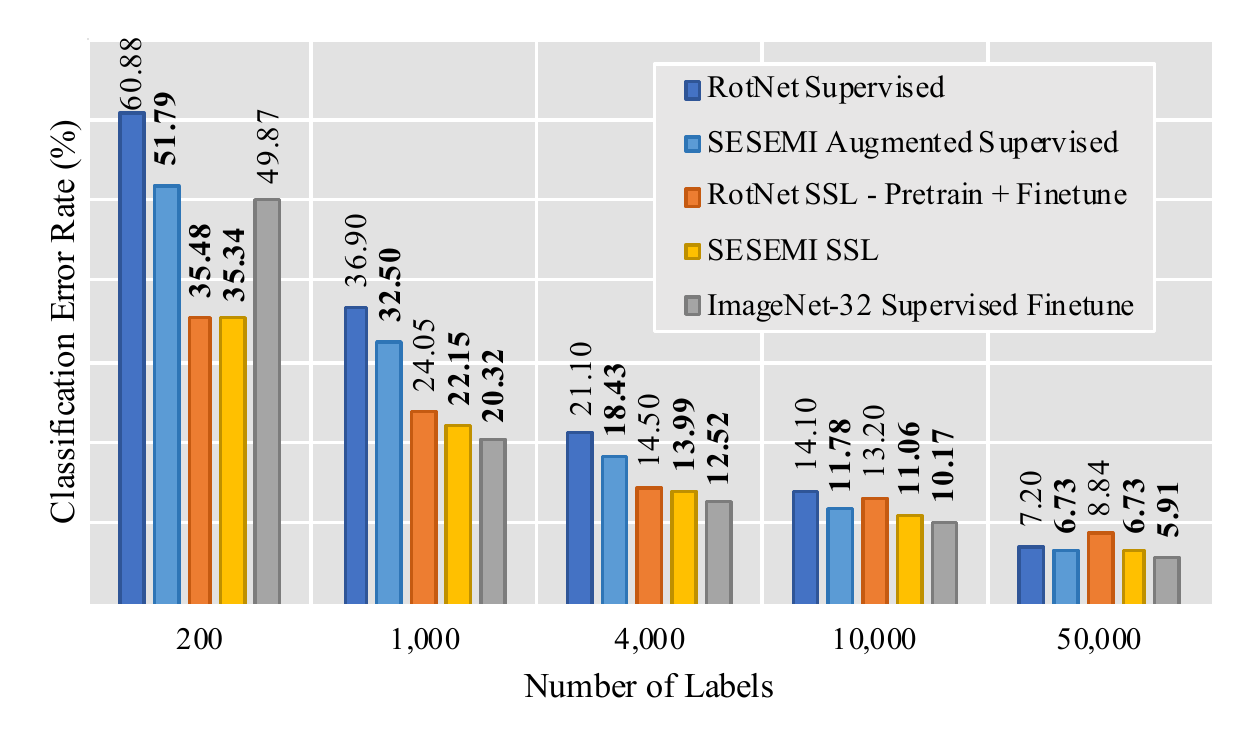}
        \caption{Comparison of SESEMI with RotNet for supervised and semi-supervised learning on CIFAR-10. RotNet models follow the conventional two-stage approach of self-supervised pre-training on unlabeled data followed by supervised fine-tuning on labeled data.}
        \label{figure2}
    \end{minipage}%
    ~~~~
    \begin{minipage}[t]{0.49\textwidth}
        \centering
        \includegraphics[width=\linewidth]{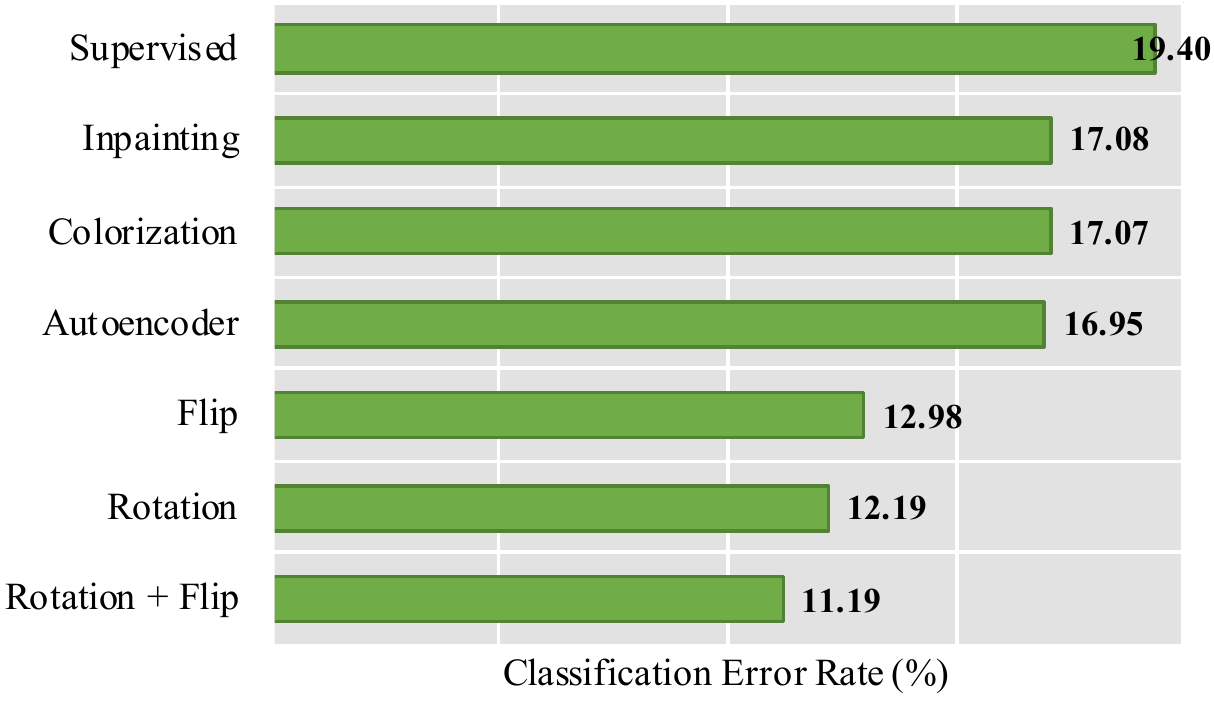}
        \caption{Exploring the relative contributions of various self-supervised tasks for SSL on CIFAR-10. The \texttt{Supervised} baseline utilizes a random sample of 4,000 labeled examples, whereas all other tasks learn from the same set of 4,000 labels along with 50,000 unlabeled examples.}
        \label{figure3}
    \end{minipage}
\end{figure}

Following the protocol of Gidaris et al. (2018) \cite{rotations}, we evaluate the accuracy of SESEMI using varying quantities of labeled examples from 200 to 50,000. \Cref{figure2} presents our supervised and semi-supervised results on CIFAR-10 with those previously obtained by RotNet. The best results are in boldface indicating the lowest classification error rate. For both supervised and semi-supervised learning, we find that our SESEMI models trained with self-supervised regularization significantly outperform RotNet models by as much as 23.9\%. Why does SESEMI outperform RotNet when the two models ostensibly share the same architecture and self-supervised task? This question can be partly explained by empirical observations that suggest \emph{better performance on the pre-training task does not always translate to higher accuracy on downstream tasks via supervised fine-tuning} \cite{revisit-ssl}. By solving both supervised and self-supervised objectives during training, SESEMI is able to learn complementary visual features from labeled and unlabeled data simultaneously for enhanced generalization.

We also perform an experiment to pre-train the NiN model on the large ImageNet-32 dataset containing 1.28 million images and then transfer to CIFAR-10 via supervised fine-tuning. Our motivation is to gain insight into the potential upper bound of supervised learning with limited labels via transfer learning. We find that our SESEMI models compare favorably to supervised transfer learning without the need to pre-train a separate model on ImageNet-scale labeled dataset. The \texttt{ImageNet-32} entry is regarded as the upper bound in performance, whereas the \texttt{Supervised} entry indicates the lower bound.

\begin{table}[t]
{\caption{Test classification error rates (\%) for supervised and semi-supervised learning on SVHN (\textbf{Left}) and CIFAR-10 (\textbf{Right}) with data augmentation averaged over four runs.}\smallskip \label{table1}}
\begin{minipage}[t]{0.5\textwidth}
\centering
\resizebox{\columnwidth}{!}{
	\begin{tabular} {lrrrrr}
	\toprule
	\multicolumn{1}{l}{\multirow{2}{*}{Method (SVHN)}} &
	\multicolumn{1}{c}{\multirow{1}{*}{~~250 labels}} &
    \multicolumn{1}{c}{\multirow{1}{*}{~~500 labels}} &
	\multicolumn{1}{c}{\multirow{1}{*}{1,000 labels}} &
	\multicolumn{1}{c}{\multirow{1}{*}{73,257 labels}} \\
    & \multicolumn{1}{c}{\multirow{1}{*}{73,257 images}}
    & \multicolumn{1}{c}{\multirow{1}{*}{73,257 images}}
    & \multicolumn{1}{c}{\multirow{1}{*}{73,257 images}}
    & \multicolumn{1}{c}{\multirow{1}{*}{~~73,257 images}} \\ 
    \midrule \midrule
    Supervised \cite{mean-teacher}
	            & $27.77 \pm 3.18$
				& $16.88 \pm 1.30$
                & $12.32 \pm 0.95$
				& $2.75 \pm 0.10$ \\
	Mixup \cite{ict}
	            & $33.73 \pm 1.79$
				& $21.08 \pm 0.61$
                & $13.70 \pm 0.47$
				& -- \\
	Manifold Mixup \cite{ict}
	            & $31.75 \pm 1.39$
				& $20.57 \pm 0.63$
                & $13.07 \pm 0.53$
				& -- \\
	SESEMI ASL (ConvNet)
	            & \bfseries 23.60 $\pm$ 1.38
				& \bfseries 15.45 $\pm$ 0.79
                & \bfseries 10.32 $\pm$ 0.16
				& \bfseries 2.26 $\pm$ 0.07 \\
	ImageNet-32 Fine-tuned
	            & $30.87 \pm 1.41$
				& $17.42 \pm 0.59$
                & $12.41 \pm 0.67$
				& $2.66 \pm 0.02$ \\
	\midrule
    $\Pi$ Model SSL \cite{tempens}
				& --
				& $6.65 \pm 0.53$
                & $4.82 \pm 0.17$
				& $2.54 \pm 0.04$ \\
	TempEns SSL \cite{tempens}
				& --
				& $5.12 \pm 0.13$
                & $4.42 \pm 0.16$
				& $2.74 \pm 0.06$ \\
    VAT SSL \cite{vat2}
    		    & --
                & --
                & $5.42 \pm 0.22$
                & -- \\
	Mean Teacher SSL \cite{mean-teacher}
    		    & \bfseries 4.35 $\pm$ 0.50
                & \bfseries 4.18 $\pm$ 0.27
                & \bfseries 3.95 $\pm$ 0.19
                & $2.50 \pm 0.05$ \\
    SESEMI SSL (ConvNet) 
                & $8.32 \pm 0.13$
                & $6.50 \pm 0.28$
                & $5.59 \pm 0.12$
                & \bfseries 2.26 $\pm$ 0.07 \\
    SESEMI SSL (WRN) 
                & $10.24 \pm 0.24$
                & $7.91 \pm 0.24$
                & $5.35 \pm 0.15$
                & \bfseries 2.34 $\pm$ 0.05 \\
    \bottomrule
	\end{tabular}
	}
\end{minipage}
\begin{minipage}[t]{0.5\textwidth}
\centering
\resizebox{\columnwidth}{!}{
	\begin{tabular} {lrrrrr}
	\toprule
	\multicolumn{1}{l}{\multirow{2}{*}{Method (CIFAR-10)}} &
	\multicolumn{1}{c}{\multirow{1}{*}{1,000 labels}} &
    \multicolumn{1}{c}{\multirow{1}{*}{2,000 labels}} &
	\multicolumn{1}{c}{\multirow{1}{*}{4,000 labels}} &
	\multicolumn{1}{c}{\multirow{1}{*}{50,000 labels}} \\
    & \multicolumn{1}{c}{\multirow{1}{*}{50,000 images}}
    & \multicolumn{1}{c}{\multirow{1}{*}{50,000 images}}
    & \multicolumn{1}{c}{\multirow{1}{*}{50,000 images}}
    & \multicolumn{1}{c}{\multirow{1}{*}{~~50,000 images}} \\
    \midrule \midrule
    Supervised \cite{mean-teacher}
	            & $46.43 \pm 1.21$
				& $33.94 \pm 0.73$
                & $20.66 \pm 0.57$
				& $5.82 \pm 0.15$ \\
	Mixup \cite{ict}
	            & $36.48 \pm 0.15$
				& $26.24 \pm 0.46$
                & $19.67 \pm 0.16$
				& -- \\
	Manifold Mixup \cite{ict}
	            & $34.58 \pm 0.37$
				& $25.12 \pm 0.52$
                & $18.59 \pm 0.18$
				& -- \\
	SESEMI ASL (ConvNet)
	            & $29.44 \pm 0.24$
				& $21.53 \pm 0.18$
                & $16.15 \pm 0.12$
				& $4.70 \pm 0.11$ \\
	ImageNet-32 Fine-tuned
	            & \bfseries 17.96 $\pm$ 0.34
				& \bfseries 12.92 $\pm$ 0.47
                & \bfseries 10.16 $\pm$ 0.22
				& \bfseries 4.61 $\pm$ 0.11 \\
	\midrule
    $\Pi$ Model SSL \cite{tempens}
				& --
				& --
                & $12.36 \pm 0.31$
				& $5.56 \pm 0.10$ \\
	TempEns SSL \cite{tempens}
				& --
				& --
                & $12.16 \pm 0.24$
				& $5.60 \pm 0.10$ \\
    VAT SSL \cite{vat2}
    		    & --
                & --
                & \bfseries 11.36 $\pm$ 0.34
                & $5.81 \pm 0.02$ \\
	Mean Teacher SSL \cite{mean-teacher}
    		    & $21.55 \pm 1.48$
                & $15.73 \pm 0.31$
                & $12.31 \pm 0.28$
                & $5.94 \pm 0.15$ \\
    SESEMI SSL (ConvNet)
                & \bfseries 17.88 $\pm$ 0.29
                & \bfseries 14.22 $\pm$ 0.27
                & \bfseries 11.65 $\pm$ 0.13
                & \bfseries 4.70 $\pm$ 0.11 \\
    SESEMI SSL (WRN)
                & \bfseries 18.32 $\pm$ 0.28
                & \bfseries 14.45 $\pm$ 0.25
                & \bfseries 11.23 $\pm$ 0.22
                & \bfseries 4.71 $\pm$ 0.23 \\
    \bottomrule
	\end{tabular}
	}
\end{minipage}
\end{table}

\begin{table}[t]
{\caption{Test classification error rates (\%) on CIFAR-100 with data augmentation averaged over four runs. \textbf{Left} -- Results with 10,000 and 50,000 labels. \textbf{Right} -- Results with unlabeled Tiny Images.}\smallskip \label{table2}}
\begin{minipage}[t]{0.5\textwidth}
\centering
\resizebox{\columnwidth}{!}{
	\begin{tabular} {lrrr}
	\toprule
	\multicolumn{1}{l}{\multirow{2}{*}{Method}} &
	\multicolumn{1}{c}{\multirow{1}{*}{~~10,000 labels}} &
	\multicolumn{1}{c}{\multirow{1}{*}{~~50,000 labels}} \\
    & \multicolumn{1}{c}{\multirow{1}{*}{~~~~50,000 images}}
    & \multicolumn{1}{c}{\multirow{1}{*}{~~~~50,000 images}} \\
    \midrule \midrule
    Supervised \cite{tempens}
                & $44.56 \pm 0.30$
				& $26.42 \pm 0.17$ \\
	SESEMI ASL (ConvNet)
                & $40.57 \pm 0.20$
				& \bfseries 22.49 $\pm$ 0.15 \\
	ImageNet-32 Fine-tuned
                & \bfseries 32.44 $\pm$ 0.27
				& \bfseries 22.22 $\pm$ 0.25 \\
	\midrule
    $\Pi$ Model SSL \cite{tempens}
				& $39.19 \pm 0.36$
				& $26.32 \pm 0.04$ \\
	TempEns SSL \cite{tempens}
				& \bfseries 38.65 $\pm$ 0.51
				& $26.30 \pm 0.15$ \\
    SESEMI SSL (ConvNet)
                & \bfseries 38.71 $\pm$ 0.11
                & \bfseries 22.49 $\pm$ 0.15 \\
    SESEMI SSL (WRN)
                & \bfseries 38.69 $\pm$ 0.10
                & $23.42 \pm 0.11$ \\
    \bottomrule
	\end{tabular}
	}
\end{minipage}
\begin{minipage}[t]{0.5\textwidth}
\centering
\resizebox{0.965\columnwidth}{!}{
	\begin{tabular} {lrrr}
	\toprule
	\multicolumn{1}{l}{\multirow{2}{*}{Method}} &
	\multicolumn{1}{r}{\multirow{1}{*}{~~~~50,000 labels}} &
	\multicolumn{1}{r}{\multirow{1}{*}{~~~~50,000 labels}} \\
    & \multicolumn{1}{r}{\multirow{1}{*}{~~~~Tiny 500,000}}
    & \multicolumn{1}{r}{\multirow{1}{*}{~~~~Tiny 237,203}} \\
    \midrule \midrule
    Supervised \cite{tempens}
                & $26.42 \pm 0.17$
				& $26.42 \pm 0.17$ \\
	SESEMI ASL (ConvNet)
                & \bfseries 22.49 $\pm$ 0.15
				& \bfseries 22.49 $\pm$ 0.15 \\
	ImageNet-32 Fine-tuned
                & \bfseries 22.22 $\pm$ 0.25
				& \bfseries 22.22 $\pm$ 0.25 \\
	\midrule
    $\Pi$ Model SSL \cite{tempens}
				& $25.79 \pm 0.17$
				& $25.43 \pm 0.32$ \\
	TempEns SSL \cite{tempens}
				& $23.62 \pm 0.23$
				& $23.79 \pm 0.24$ \\
    SESEMI SSL (ConvNet)
                & \bfseries 22.52 $\pm$ 0.10
                & \bfseries 22.50 $\pm$ 0.26 \\
    SESEMI SSL (WRN)
                & \bfseries 22.65 $\pm$ 0.30
                & \bfseries 22.62 $\pm$ 0.24 \\
    \bottomrule
	\end{tabular}
	}
\end{minipage}
\end{table}

\subsubsection{Self-Supervised Regularization Outperforms Consistency Regularization on CIFAR-10 and CIFAR-100}

\textbf{SVHN} ~ \Cref{table1} compares our supervised and semi-supervised results with consistency baselines. In analyzing the SVHN results on the left side of \Cref{table1}, we observe that SESEMI ASL surpasses the supervised baselines, including \texttt{ImageNet-32} and those with strong Mixup \cite{mixup} and Manifold Mixup \cite{manifold-mixup} regularization, by a large margin for all experiments. However, the results are not satisfactory when compared against the semi-supervised baselines, especially Mean Teacher. We discuss the limitation of SESEMI for semi-supervised learning on the SVHN dataset in \Cref{discuss}.

\textbf{CIFAR-10} ~ Experiments on CIFAR-10 tell a different story. The right side of \Cref{table1} shows that SESEMI uniformly outperforms all supervised and semi-supervised baselines, improving on SSL results by as much as 17\%. On CIFAR-10, the combination of supervised and self-supervised learning is a strength of SESEMI, but it is also a limitation in the case of SVHN. We observe that the ConvNet and WRN architectures produce comparable results across the board.

\textbf{CIFAR-100 and Tiny Images} ~ The successes of SESEMI on CIFAR-10 also transfer to experiments on CIFAR-100. The left side of \Cref{table2} provides a comparison of SESEMI against the $\Pi$ model and TempEns baselines, where we obtain competitive semi-supervised performance using 10,000 labels and achieve state-of-the-art supervised results when all 50,000 labels are available, matching the upper bound performance of ImageNet-32 supervised fine-tuning.

Additionally, we run two experiments to evaluate the performance of SESEMI in the case of class-distribution mismatch. Following Laine and Aila (2017) \cite{tempens}, our first experiment utilizes all 50,000 available labels from CIFAR-100 and randomly samples 500,000 unlabeled extra Tiny Images, most belonging to categories not found in CIFAR-100. Our second experiment uses a restricted set of 237,203 Tiny Images from categories found in CIFAR-100. The right side of \Cref{table2} presents SSL error rates on CIFAR-100 augmented with Tiny Images. Results show that adding 500,000 unlabeled extra data with significant class-distribution mismatch does not degrade SESEMI performance, as observed by Oliver et al. (2018) \cite{ssl-eval} with other SSL approaches. For SESEMI with WRN-28-2 architecture, the addition of (randomly selected) unlabeled extra data from Tiny Images further reduces CIFAR-100 error rate from 23.42\% to 22.62\%, matching performances obtained by SESEMI with ConvNet and ImageNet-32 supervised fine-tuning.

\begin{table}[t]
{\caption{Test classification error rates (\%) on CIFAR-10 with 4,000 labels and SVHN with 1,000 labels. All entries are trained using the WRN-28-2 architecture within the unified evaluation framework of Oliver et al. (2018) \cite{ssl-eval}.}\label{table3}}
\begin{center}
\begin{adjustbox}{width=\textwidth}
	\begin{tabular} {lrrrrrrr}
	\toprule
	\multicolumn{1}{l}{\multirow{1}{*}{Dataset}} &
	\multicolumn{1}{l}{\multirow{1}{*}{\# Labels}} &
	\multicolumn{1}{c}{\multirow{1}{*}{Supervised}} &
    \multicolumn{1}{c}{\multirow{1}{*}{$\Pi$ Model}} &
	\multicolumn{1}{c}{\multirow{1}{*}{Mean Teacher}} &
	\multicolumn{1}{c}{\multirow{1}{*}{VAT}} & 
	\multicolumn{1}{c}{\multirow{1}{*}{ImageNet32}} & 
	\multicolumn{1}{c}{\multirow{1}{*}{SESEMI (ours)}} \\
    \midrule \midrule
    CIFAR-10
				& 4,000
				& $20.26 \pm 0.38$
                & $16.37 \pm 0.63$
				& $15.87 \pm 0.28$
				& $13.86 \pm 0.27$
				& $12.09$
				& \bfseries 11.23 $\pm$ 0.22 \\
    SVHN
				& 1,000
				& $12.83 \pm 0.47$
                & $7.19 \pm 0.27$
				& $5.65 \pm 0.47$
				& $5.63 \pm 0.20$
				& --
				& \bfseries 5.35 $\pm$ 0.15 \\
    \bottomrule
	\end{tabular}
\end{adjustbox}
\end{center}
\end{table}

\begin{table}
\centering
{\caption{Test classification error rates (\%) for various supervised methods on CIFAR-10 and CIFAR-100 with all 50,000 labeled training examples. For each method, we compare best-case performances along with the required number of trainable parameters.}\smallskip \label{table4}}
\resizebox{0.65\columnwidth}{!}{
    \begin{tabular} {lrrrr}
	\toprule
	Method & 
	\# Params &
    CIFAR-10 &
	CIFAR-100 \\
    \midrule \midrule
    FractalNet \cite{fractalnet}
				& 38.6M
				& 4.60
                & 23.73 \\
    Fractional MP \cite{fmp}
	            & 12.0M
				& 3.47
                & 26.39 \\
    ResNet-1001 \cite{resnet1001}
				& 10.2M
				& 4.62
                & 22.71 \\
	Wide ResNet-40-4 \cite{wrn}
				& 8.9M
				& 4.53
                & 21.18 \\
    DenseNet ($k = 12$) \cite{densenet}
				& 7.0M
				& 4.10
                & 20.20 \\
    FitResNet (LSUV) \cite{fitresnet}
				& 2.5M
				& 5.84
                & 27.66 \\
    Highway Network \cite{highway}
				& 2.3M
				& 7.72
                & 32.39 \\
    \midrule
    Supervised (ConvNet) \cite{mean-teacher,tempens}
				& 3.1M
				& 5.82
                & 26.42 \\
    SESEMI ASL (ConvNet)
				& 3.1M
				& 4.70
                & 22.49 \\
    SESEMI ASL (WRN-28-2)
				& 1.5M
				& 4.71
                & 23.42 \\
    \bottomrule
	\end{tabular}
	}
\end{table}

\textbf{SESEMI with Residual Networks} ~ \Cref{table3} provides a comparison of SESEMI for semi-supervised learning on SVHN and CIFAR-10 within the unified evaluation framework of Oliver et al. (2018) \cite{ssl-eval}, in which they re-implemented the consistency baselines using the WRN-28-2 architecture, carried out large-scale hyper-parameter optimization specific to each technique, and reported best-case performances. Our SESEMI model with WRN-28-2 architecture establishes a new upper bound in SSL performance by outperforming all methods, including \texttt{ImageNet-32}, under this evaluation setting for both SVHN and CIFAR-10.

It is important to note that we do not perform any hyper-parameter search in this experiment, but use the same set of hyper-parameters described in \Cref{implementation} along with $w = 1$ for the weighting of the self-supervised loss term in SESEMI. In practical applications where tuning many (possibly inter-dependent) hyper-parameters can be problematic \cite{ssl-eval}, especially over small validation sets, our approach to supervised and semi-supervised learning using SESEMI offers a clear and significant benefit.

\subsection{Comparison with State-of-the-Art Supervised Methods on CIFAR-10 and CIFAR-100}
Motivated by the strong performances of SESEMI ASL for supervised learning \emph{augmented} with self-supervised regularization, we provide a comparative analysis of SESEMI against several previous state-of-the-art supervised methods on CIFAR-10 and CIFAR-100 in \Cref{table4}. We observe that SESEMI ASL is competitive in predictive performance with advanced CNN architectures like FractalNet \cite{fractalnet}, Fractional Max-Pooling \cite{fmp}, ResNet-1001 \cite{resnet1001}, Wide ResNet-40-4 \cite{wrn}, and DenseNet \cite{densenet} while requiring a fraction of the computational complexity, as measured in millions of trainable parameters. For those architectures having roughly the same number of trainable parameters, our SESEMI ASL models outperform Highway Network \cite{highway} and FitResNet with LSUV initialization \cite{fitresnet} by a large margin.

The effectiveness of SESEMI ASL is directly attributed to self-supervised regularization and not to the CNN architecture. The same ConvNet architecture without self-supervised regularization performs significantly worse on both CIFAR-10 (5.82\% error rate) and CIFAR-100 (26.42\% error rate). In principle, self-supervised regularization could be incorporated into any CNN architecture for further reduction in classification error rate.

\subsection{Ablation Study}
Several studies have provided conclusive evidence that self-supervision is an effective unsupervised pre-training technique for downstream supervised visual understanding tasks such as image classification, object detection, and semantic segmentation \cite{multiself,revisit-ssl}. However, the evaluation of self-supervised algorithms for SSL has not been explored, especially in the setting where the supervised and self-supervised losses are jointly trained, per Algorithm~\ref{algorithm1}. We briefly describe the following self-supervised tasks based on image reconstruction and compare their SSL performances against the task of classifying image rotations and flips. All experiments use the same convolutional encoder-decoder framework, where the encoder backbone is the WRN-28-2 architecture, and the decoder head comprises a set of two deconvolutional layers \cite{fcn} with batch normalization and ReLU non-linearity to produce a reconstructed output with the same dimensions as the input.

\textbf{Denoising Autoencoder} ~ The self-supervised objective of this simple baseline is to minimize the mean pixel-wise squared error between the reconstructed output and image input corrupted with Gaussian noise.

\textbf{Image Inpainting} ~ Following \cite{inpainting}, the input to the encoder is an image with the central square patch covering \nicefrac{1}{4} of the image masked out or set to zero. The decoder is trained to generate prediction for the masked region using a masked $L_2$ reconstruction loss as self-supervision.

\textbf{Image Colorization} ~ Following \cite{colorization}, the input to the encoder is a grayscale image (the L* channel of the L*a*b* color space) and the decoder is trained to predict the a*b* color components at every pixel. The self-supervised loss is the mean squared error between the reconstructed a*b* color output and ground truth a*b* color components.

\textbf{Ablation Results} ~ \Cref{figure3} shows the individual tasks of recognizing image flips and rotations outperform image reconstruction, inpainting, and colorization on the CIFAR-10 dataset. These results suggest that classification-based self-supervision provides a better, or perhaps more compatible, proxy label for semi-supervised image classification than reconstruction-based tasks. Our findings corroborate recent studies showing rotation-based self-supervision is the superior pre-training technique for downstream transfer learning tasks \cite{rotations,revisit-ssl}. Lastly, combining horizontal and vertical flips with the rotation recognition task outperforms all other self-supervised tasks, leading to an improvement in SSL performance over rotation recognition by 8.2\%.

\section{Discussion}\label{discuss}
\textbf{Limitation of SESEMI} ~ We speculate the poor performance of SESEMI on the SVHN dataset stems from our chosen self-supervised task of predicting image rotations and flips. Gidaris et al. (2018) \cite{rotations} showed that their self-supervised model focused its attention maps on salient parts of the images to aid in the rotation recognition task. We hypothesize similar dynamics are at play here, but the SVHN dataset presents an additional layer of complexity in which the centermost digits (the digits to be recognized) are often surrounded by ``distractor'' digits. When the digits are rotated and flipped, the self-supervised branch is likely picking up dominant visual features corresponding to the distractor digits and relate them to the supervised branch as belonging to the digits of interest. These ``miscues'' are most prominent when few labels are present, where the supervised branch is simply learning visual information from the self-supervised branch. However, when all labels are available, the supervised branch is able to correct the miscues, and our SESEMI models produce the best classification results.

\textbf{Comparison with Recent Developments in SSL} ~ Recent years have seen a flurry of research on SSL techniques that are related to or concurrent with our work. The prior work of Smooth Neighbors on Teacher Graphs \cite{sntg}, Virtual Adversarial Dropout \cite{vadd}, Interpolation Consistency Training \cite{ict}, Stochastic Weight Averaging \cite{swa}, and Label Propagation \cite{label-prop} advanced the field of semi-supervised learning by achieving impressive results on SVHN, CIFAR-10, and CIFAR-100 benchmarks. However, those methods all build upon strong consistency baselines by either adding a third loss (and new hyper-parameters to tune) to the overall objective of consistency models or averaging model weights. The concurrent work on self-supervised semi-supervised learning \cite{s4l} independently explores the contributions of self-supervised regularization for SSL in a way similar to ours, but with a different evaluation protocol and end goal. MixMatch \cite{mixmatch} combines strong Mixup \cite{mixup} regularization with consistency regularization to achieve state-of-the-art SSL results.

Our goal for this work is to directly compare the effectiveness of self-supervised regularization to consistency regularization, which has not been done before. Moreover, we deliberately avoid increasing complexity in favor of simplicity and pragmatism with our design choices and training protocol. In principle, our work is potentially complementary to Label Propagation and MixMatch. Label Propagation requires the Mean Teacher prediction function to work well, which can be directly replaced by our SESEMI module. For MixMatch, we can integrate a third loss term into the objective function for learning additional self-supervised features. These are viable topics for future research.

\section{Conclusion}
We presented a conceptually simple yet effective multi-task CNN architecture for supervised and semi-supervised learning based on self-supervised regularization. Our approach produces proxy labels from geometric transformations on unlabeled data, which are combined with ground truth labels for improved learning and generalization in the limited labeled data setting. We provided a comprehensive empirical evaluation of our approach using three different CNN architectures, spanning multiple benchmark datasets and baselines to demonstrate its effectiveness and wide range of applicability. We highlight two attractive benefits of SESEMI. First, SESEMI achieves state-of-the-art predictive performance for both supervised and semi-supervised image classification without introducing additional hyper-parameters to tune. Second, SESEMI does not need a separate pre-training step, but is trained end-to-end for simplicity, efficiency, and practicality.

\section*{Acknowledgments}
The author thanks Cole Winans and Brian Keller at Flyreel for their continued support, and gracious reviewers for their constructive feedback on this paper.

\bibliographystyle{ecai}
\bibliography{refs}

\newpage

\section*{Appendix}

\appendix

\section{Dataset Details}\label{datasets}
The SVHN \cite{svhn} dataset contains 73,257 train and 26,032 test samples categorized over 10 digits (0-9) in natural scene images. The classification task is to recognize the centermost digit in each color image of $32 \times 32$ pixels. We only use the official train/test splits and do not utilize the provided 531,131 extra images. The CIFAR-10 \cite{cifar} dataset consists of 60,000 $32 \times 32$ natural color images in 10 classes, with 6,000 images per class. The dataset is split into 50,000 train and 10,000 test samples. The CIFAR-100 \cite{cifar} dataset is similar to CIFAR-10, except it has 100 classes containing 600 images each. There are 500 train and 100 test images per class. CIFAR-10 and CIFAR-100 are labeled subsets of the 80 million Tiny Images dataset \cite{tiny-images}, which is organized into 75,062 generic scene and object categories. ImageNet-32 \cite{tiny-imagenet} is the full ImageNet dataset \cite{imagenet} down-sampled to $32 \times 32$ pixels. We use ImageNet-32 for supervised transfer learning experiments. We use Tiny Images as a source of unlabeled extra data to augment semi-supervised learning on CIFAR-100 and to evaluate the performance of SESEMI under the condition of class-distribution mismatch.

\section{Implementation Details}\label{implementation}
We implement the SESEMI algorithm using Keras \cite{keras} with GPU-enabled TensorFlow backend \cite{tf}. We follow standard practice for data pre-processing, augmentation, and hyper-parameter search.

\textbf{Data Pre-processing and Augmentation} ~ We apply global contrast normalization to scale all datasets to have zero mean and unit $L_2$ norm. We further pre-process CIFAR-10, CIFAR-100, and Tiny Images with Zero Components Analysis (ZCA) whitening \cite{cifar}. Standard data augmentation on CIFAR-10 and CIFAR-100 includes random translations by up to 2 pixels on each side $\{\Delta x, \Delta y\} \in [-2, 2]$, horizontal (left-right) flip, and additive Gaussian noise with $\sigma=0.15$, whereas SVHN is limited to random translations and Gaussian noise. Data augmentation is applied independently to both supervised and self-supervised branches of SESEMI, except we do not apply horizontal flip on the self-supervised branch.

\textbf{Hyper-parameters} ~ During model development, we use 10 percent of the provided SVHN training examples as a \emph{dev} set and perform hyper-parameter tuning to find the optimal combination of mini-batch size, percentage of dropout regularization, initial learning rate, and number of training epochs that minimizes classification error on said dev set. The same hyper-parameters are subsequently shared across supervised and semi-supervised settings for all datasets, and are used in all experiments featuring NiN, ConvNet and WRN architectures. These are standard hyper-parameters for tuning CNNs and are not specific to SESEMI, which is a notable advantage of our approach. By contrast, previous consistency baselines have specific hyper-parameters that must be carefully tuned for optimal performance, such as the consistency coefficient in $\Pi$ model, exponential moving average decay in Mean Teacher, and norm constraint $\epsilon$ for the adversarial direction in VAT. The final models are trained on all examples from the combined training and dev sets.

\textbf{Training Protocol} ~ We train our models using Nesterov accelerated gradient descent \cite{nag} on mini-batches of 16 examples, with an initial base learning rate of 0.05, momentum of 0.9, weight decay of 0.0005, and dropout rate of 0.5 in all experiments. Similar to \cite{rotations}, we implement four rotations and two flips on a given image in a mini-batch for improved training. Thus, our SESEMI models receive two effective mini-batches having the same number of $16 \times 6 = 96$ unlabeled and labeled examples. In the supervised setting, we train SESEMI for up to 2,000 epochs over labeled examples in $\mathcal{D}_L$, depending on the dataset and scarcity of labeled examples. In the semi-supervised setting, we train SESEMI over unlabeled examples in $\mathcal{D}_U$ for 30 epochs on SVHN and 50 epochs on CIFAR-10 and CIFAR-100.

During training, we anneal the base learning rate according to the polynomial decay of the form: $\text{lr}(t) \leftarrow \text{base\_lr} \times \left(1 - \nicefrac{t}{t_\text{max}}\right)^p$, where $\text{base\_lr} = 0.05$, $t$ is the current iteration, $t_\text{max}$ is the maximum number of iterations, and $p = 0.5$ controls the rate of decay. With this learning rate schedule, the models gain the most performance improvement in the last few epochs, so we simply report test error after the last training iteration. We report the mean and standard deviation of four independent runs with random weight initializations on fixed data splits.

\end{document}